\documentclass[10pt,twocolumn,letterpaper]{article}

\usepackage[pagenumbers]{cvpr} 

\usepackage[dvipsnames]{xcolor}

\definecolor{cvprblue}{rgb}{0.21,0.49,0.74}
\usepackage[pagebackref,breaklinks,colorlinks,citecolor=cvprblue]{hyperref}

\usepackage[ruled]{algorithm2e}
\usepackage{amsmath}
\usepackage{makecell}

\def\paperID{9763} 
\def\confName{CVPR}
\def\confYear{2024}

\title{Lightweight Diffusion Models with Distillation-Based Block Neural Architecture Search}

\author{Siao Tang\\
Tsinghua University\\
\and
Xin Wang \thanks{Corresponding author}   \\
Tsinghua University \\
\and
Hong Chen \\
Tsinghua University \\
\and
Chaoyu Guan \\
Tsinghua University \\
\and
Yansong Tang \\
Tsinghua University \\
\and
Wenwu Zhu *   \\
Tsinghua University \\
}

\begin{document}
\maketitle
\begin{abstract}
Diffusion models have recently shown remarkable generation ability, achieving state-of-the-art performance in many tasks. However, the high computational cost is still a troubling problem for diffusion models. 
To tackle this problem, we propose to automatically remove the structural redundancy in diffusion models with our proposed Diffusion Distillation-based Block-wise Neural Architecture Search (DiffNAS). Specifically, given a larger pretrained teacher, we leverage DiffNAS to search for the smallest architecture which can achieve on-par or even better performance than the teacher. Considering current diffusion models are based on UNet which naturally has a block-wise structure, we perform neural architecture search independently in each block, which largely reduces the search space. Different from previous block-wise NAS methods, DiffNAS contains a block-wise local search strategy and a retraining strategy with a joint dynamic loss. Concretely, during the search process, we block-wisely select the best subnet to avoid the unfairness brought by the global search strategy used in previous works. When retraining the searched architecture, we adopt a dynamic joint loss to maintain the consistency between supernet training and subnet retraining, which also provides informative objectives for each block and shortens the paths of gradient propagation. We demonstrate this joint loss can effectively improve model performance. We also prove the necessity of the dynamic adjustment of this loss. The experiments show that our method can achieve significant computational reduction, especially on latent diffusion models with about 50\% MACs and Parameter reduction.

\end{abstract}
    
\section{Introduction}
The diffusion model is one of the generative models which learns the underlying data distribution by adding noise and iteratively denoising~\cite{ho2020denoising,song2020denoising,liupseudo}. Diffusion models have shown remarkable generative ability, outperforming advanced GANs and VAEs on a series of benchmarks~\cite{ho2022cascaded,dhariwal2021diffusion}.
Diffusion models have been applied to a variety of tasks and achieved great success, such as text-to-image generation~\cite{rombach2022high}, text-to-video generation~\cite{wu2022tune}, graph generation~\cite{vignac2022digress}, and speech synthesis~\cite{chen2021wavegrad}. 


However, a critical limitation of diffusion models is their high computational cost, which is mainly from two aspects. On the one hand, diffusion models need a long sequence of noise prediction steps (e.g., 1000 steps for DDPM~\cite{ho2020denoising}) to generate a batch of high-quality outputs. On the other hand, the computation amount of each noise prediction is large because of the high computational overhead of the network architecture. To tackle the first problem, substantial works~\cite{liupseudo,ludpm,salimansprogressive,luhman2021knowledge,watson2021learning,lyu2022accelerating} engage in reducing sampling steps by finding more effective sampling trajectories or knowledge distillation.


\begin{figure}
    \centering
    \includegraphics[width=0.47\textwidth]{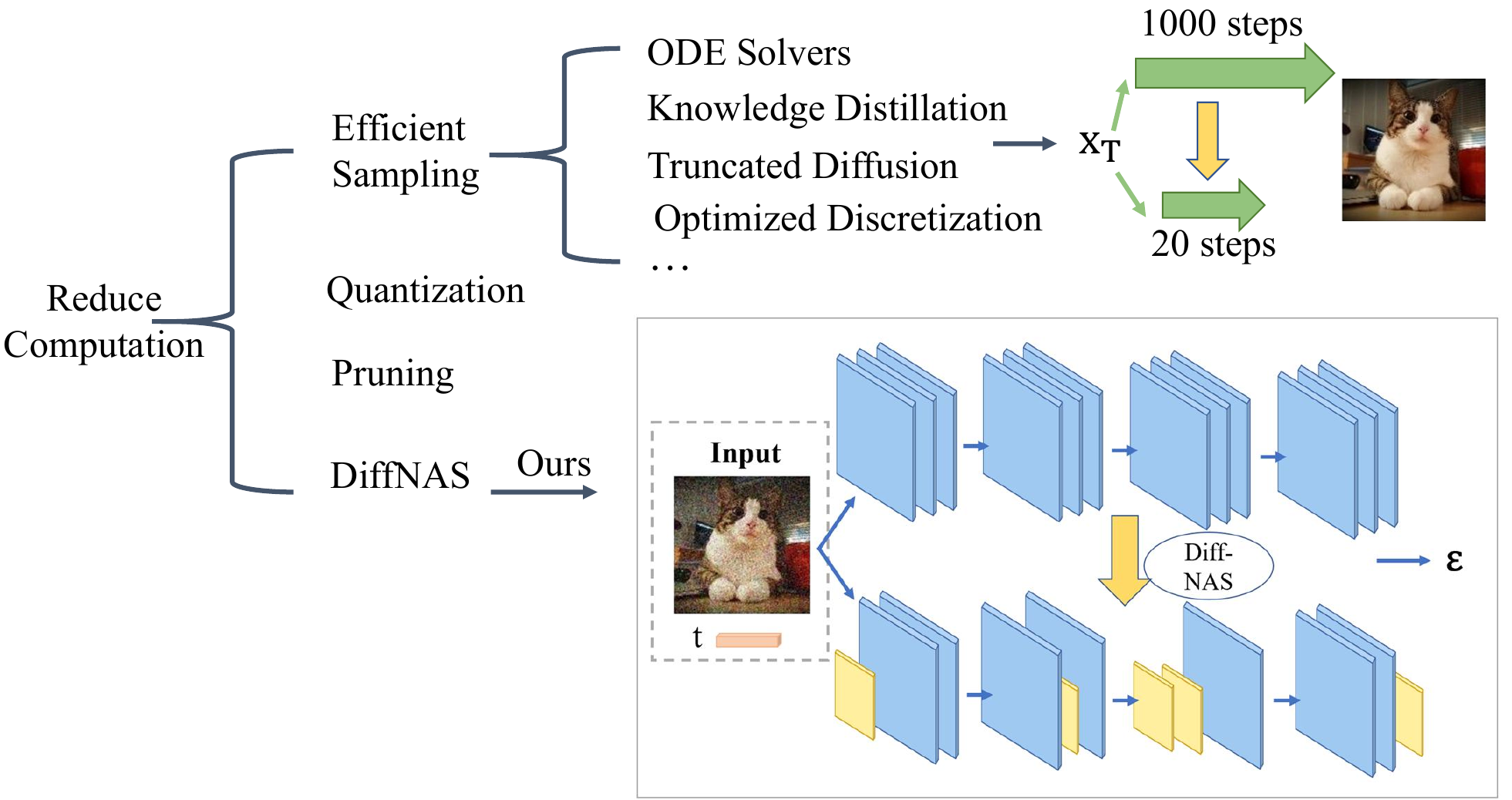}
    \caption{ Concept comparison between DiffNAS and previous computation reduction works.
    Previous works mainly focus on efficient sampling or quantization. From another important perspective, we propose to remove the structural redundancy of diffusion models by automatically searching for the most efficient architecture.}
    \label{fig: taxonomy}
\end{figure}

We focus on the second problem, i.e., the high computational overhead brought by large network architecture.
Quantization~\cite{shang2023post,li2023q} and pruning methods~\cite{fang2023structural} have been explored to compress diffusion models.
From an orthogonal perspective, we engage in removing the structural redundancy for diffusion models by automatically searching for the most efficient architecture.
Figure~\ref{fig: taxonomy} shows the concept comparison between our method and previous works.
Specifically, we propose a Diffusion Distillation-based Block-wise Neural Architecture Search (DiffNAS) framework, which searches for the smallest architecture whose generative quality is better or on par with a pretrained teacher, through the proposed block-wise local search strategy during NAS evaluation and the proposed dynamic joint loss during subnet retraining. Our method is orthogonal to quantization or pruning methods and can be combined with them to achieve further compression. Furthermore, we use the performance of the baseline model (that needs to be compressed) as the constraint for our method. This constraint guarantees the obtained architecture can theoretically have better performance than the baseline, which previous quantization or pruning methods can't ensure.

Concretely, considering the pyramidal block-wise architecture of U-Net~\cite{ronneberger2015u} that most diffusion models~\cite{ho2020denoising,song2020denoising,rombach2022high,song2020improved,nichol2021improved,songscore} are based on, we adopt the block-wise NAS paradigm~\cite{li2020block,li2021bossnas} and regard each resolution level as a block. We use the baseline model expected to be compressed as the teacher.
We use the intermediate distillation loss, i.e., $L2$ loss, to supervise the block-wise training of the supernet. The loss measures the distance between the teacher output and the student output at each block. 

During the NAS evaluation, we use the distillation loss to measure the distance of subnets to the teacher, since a smaller loss means a better ability to imitate the teacher. Previous works~\cite{li2020block} use a global search strategy to search for the expected architecture under certain constraints. However, we demonstrate the global search strategy leads to unfairness across different blocks, because it fails to balance the loss value magnitude of these blocks.
Therefore, we propose to block-wisely select the smallest subnets which have a lower evaluation loss compared to the subnet having the same architecture as the teacher. This loss constraint theoretically guarantees the searched subnet can have better performance than the teacher.
Finally, we obtain the complete subnet by concatenating the searched architectures of each block.

During the retraining, we observe that retraining the searched subnet only with the original noise prediction loss leads to poor performance than the teacher. One main reason is that only using the original loss can not maintain the consistency between supernet training and subnet retraining.
Specifically, the distillation loss actually reflects the ability of these subnets to imitate the teacher when training supernet and evaluating. 
This indicates we also need the teacher to guide the retraining process of subnets.
Therefore, instead of retraining the subnet without distillation as previous works~\cite{li2020block}, we propose to train with a dynamic joint loss combining the distillation loss and the original noise prediction loss. Specifically, we gradually increase the weight of the noise prediction loss and decrease the weight of the distillation loss. 
The early retraining can be regarded as a warm-up stage when the student quickly learns the knowledge from the teacher. In the later retraining, the student needs to be refined, so the noise prediction loss becomes more important since it directly determines the accuracy of noise prediction. 
The distillation loss maintains the consistency between supernet training and retraining. Moreover, it further improves model performance since the intermediate feature maps provide information at different resolution levels and shorten the paths of gradient propagation.
We conduct experiments on various datasets with different resolutions and computation amounts. 


In conclusion, our contributions are summarized as follows:

\begin{enumerate}
    \item To compress diffusion models, we propose to remove the structural redundancy by automatically searching for the most efficient architecture.
    \item We propose DiffNAS, a framework to compress diffusion models without degrading performance. DiffNAS contains a novel block-wise search strategy to avoid unfairness brought by global search, and a novel dynamic joint loss for NAS retraining to maintain the consistency between supernet training and subnet retraining.
    
    \item The results show that our method achieves about 30\% MACs reduction on pixel-wise models, as well as about 50\% MACs and Parameters reduction on latent diffusion models, while maintaining on-par or even achieving better performance.
    
\end{enumerate}

\section{Related Work}
 
\textbf{Neural Architecture Search (NAS).} Neural Architecture Search (NAS) aims to explore more efficient architectures than manually designed ones, replacing the efforts of human experts to design models. NAS has three essentials, i.e., search space, search strategy, and evaluation strategy. Early works~\cite{zhong2018practical,zoph2016neural,bakerdesigning} sample an architecture candidate by RNN or evolutionary algorithm and evaluate its performance after training from scratch. The type of NAS costs a large amount of computation and is thus impractical. Afterward, most works employ a weight-sharing supernet to represent search space. 
Gradient-based methods~\cite{wu2019fbnet,liu2018darts} assign choosing weights to architecture options and optimize these weights with the net parameters jointly. 
One-shot methods disentangle supernet training with architecture search, usually training the supernet with path sampling or path dropout. when evaluating, subnets directly inherit parameters from the trained supernet. To reduce search space and improve the accuracy of model ranking, block-wise NAS methods are proposed~\cite{li2020block,li2021bossnas}. They use a distance loss between the feature maps of the teacher and the student
to guide block-wise supernet training and evaluation. In our work, we simulate the same supernet training process as previous works~\cite{li2020block}. However, to cope with the unfairness and the inconsistency problems, our method proposes a novel block-wise search strategy and a retraining strategy with a dynamic joint loss.

\noindent \textbf{Model Compression.} Model compression is a significant technology to lighten large neural networks. Common methods include Network Pruning, Knowledge Distillation (KD), and Quantization. Pruning aims to remove the relatively less important parameters. 
Knowledge distillation aims to leverage the knowledge of a teacher network to guide the training of a more compact student network.
Quantization refers to the process of using fewer bits to represent float point tensors and perform model inference. 
In practice, pruning might be limited by the support of inference platforms and hardware. Quantization can not cope with the redundancy from architecture.
Different from the perspective of using few bits or cutting off unimportant weights, our method engages in searching for appropriately small architectures. Therefore, our method is orthogonal to quantization and pruning, which means it can also be combined with pruning or quantization to further compress diffusion models.

\section{Preliminaries}

\textbf{Diffusion Models.} Diffusion models are probabilistic generative models that learn the underlying data distribution by adding noise to the input and gradually denoising the noisy variable.
The initial idea of diffusion models can trace back to \cite{sohl2015deep}. In recent years, diffusion models have shown powerful performance in generative tasks and attracted a significant amount of attention. Besides noise prediction, diffusion models have several variants, e.g., data prediction~\cite{ramesh2022hierarchical} and velocity prediction~\cite{ho2022imagen} models. In this paper, we discuss the most frequently-used form, i.e., noise prediction models.

Diffusion models define a forward process that adds noise to the original input:
{
\begin{align}
q\left(\boldsymbol{x}_{t} | \boldsymbol{x}_{0}\right)=\mathcal{N}\left( \alpha_t \boldsymbol{x}_{0}, \sigma^{2}_t \boldsymbol{I}\right) , \nonumber
\end{align}
}

\noindent where $\alpha_t$ and $\sigma_t$ are differentiable functions of $t$, which are determined by certain noise schedule. The formula can be reparametered as $\boldsymbol{x}_{t} = \alpha_t \boldsymbol{x}_{0} +  \sigma_t \boldsymbol{\epsilon}$, where $\boldsymbol{\epsilon} \sim \mathcal{N}(\mathbf{0}, \mathbf{I})$.
The signal-to-noise-ratio (SNR) $\alpha_t / \sigma_t$ is strictly decreasing with respect to $t$, such that the $q(x_t|x_0)$ is nearly a unit Gaussian when $t$ is large enough. The neural network $\boldsymbol{\epsilon_{\theta}}$ is trained to predict the true noise $\epsilon$ given $\boldsymbol{x}_{t}$ and $t$ , i.e., minimizing:

\begin{align}
\mathbf{E}_{\boldsymbol{x}_{0},\boldsymbol{\epsilon}, t} 
[w_{t}\left\|\boldsymbol{\epsilon}_{\theta}\left(\alpha_{t} \boldsymbol{x}_{0}+\sigma_{t} \boldsymbol{\epsilon},t \right)-\boldsymbol{\epsilon}\right\|_{2}^{2}]  . \nonumber
\end{align}

\noindent At inference time, diffusion models generate new samples by iteratively denoising $\boldsymbol{x_T} \sim \mathcal{N}(\mathbf{0}, \mathbf{I})$ using some kind of sampling strategies, e.g., DDIM~\cite{song2020denoising}, PNDM~\cite{liupseudo} and DPM-Solver~\cite{ludpm}. 

\noindent \textbf{Block-wise NAS.} Block-wise NAS is firstly proposed by Liu et al.~\cite{li2020block}. The motivation is to reduce search space and promote the accuracy of model ranking by
narrowing the performance gap between proxy subnets with shared weights and the retrained stand-alone ones. The gap means the ranks of subnets inheriting weights from the supernet are not consistent with the ranks of the retrained ones. Some works~\cite{chu2021fairnas,li2020improving} show the gap is narrowed as the search space decreases. Let $N$ denote the whole supernet, block-wise NAS divides $N$ into $n$ blocks: 
{
\begin{align}
\mathcal{N} =\mathcal{N}_{n-1}\ldots\mathcal{N}_{i+1}\circ\mathcal{N}_i\cdots\circ\mathcal{N}_0 . \nonumber
\label{eq_6}
\end{align}
}

\noindent Then, block-wise NAS train each block of the supernet separately:
{
\begin{align}
\mathcal{W}_{i}^{*}=& \min _{\mathcal{W}_{i}} \mathcal{L}_{\text {train }}\left(\mathcal{W}_{i}, \mathcal{A}_{i} ; \mathbf{X_i}, \mathbf{Y_i}\right), \quad i=0,1 \cdots n-1 , \nonumber
\end{align}
}

\noindent where $ \mathcal{A}_{i}$ denotes the search space of the $i$-th block and $\mathcal{W}_{i}$ denotes the parameters. $\mathbf{X_i}$ denotes the input of the $i$-th block, which is the output of the $(i-1)$-th block of the teacher. $\mathbf{Y_i}$ denotes the groundtruth of the $i$-th block, which is the output of the $i$-th block of the teacher. Block-wise NAS can extremely reduce the search space, from $\prod_{i=0}^{n-1} C^{d_i}$ to $C^{d_i}$, where $C$ denotes the number of the optional operations in each layer and $d_i$ denotes the number of layers in the $i$-th block. 
\section{Methodology}
Figure~\ref{framework} illustrates the whole framework of our method. Given a target pretrained model that is expected to be lightened, we first block-wisely train a supernet containing a weight-sharing search space, with distillation as the supervision. The search space includes the architecture of the teacher. Then we block-wisely search for the smallest subnet with better performance than the teacher. Finally, we retrain the subnet with a dynamic joint loss combining the knowledge distillation loss and the original loss. In this section, we will describe the three stages in detail. 
The time and memory costs of our method are not more than training a baseline, which we show details in the Appendix (\cref{sec: cost analysis}).

\subsection{Block-wise NAS Training}
Most diffusion models use UNet-based architecture which has several levels of resolutions for feature encoding and decoding. 
The encoder extracts features at different abstraction levels, from low-level textures to high-level semantics, while the decoder gradually recovers the information at different levels. Considering this naturally block-wise architecture, we divide the whole UNet into blocks, regarding each resolution level as a block. In other words, we perform NAS independently for each block. 
we use the $L2$ loss between the block output of teacher and student to supervise NAS training for each block. Let $T=\{E^t_i,D^t_i\}$ and $\ S=\{E^s_i,D^s_i\},i=0,1,...,n-1$ respectively denote the teacher and the student with $n$ resolution levels, where $E_i$ and $D_i$ denote the encoder and the decoder at $i$-th resolution level. For simplicity, we ignore the notation of the middle block that is between the encoding stage and the decoding stage.
The training objective for each block is written below:

{
\begin{align}
\mathcal{W}_{i}^{*}= \min _{\mathcal{W}_{i}} &\mathcal{L}_{\text {train }}\left(\mathcal{W}_{i}, \mathcal{A}_{i} ; \mathbf{X_i}, \mathbf{Y_i}\right), \quad i=0,1 \cdots 2n-1 , \nonumber \\ 
& \mathcal{L}_{\text {train}} = 
||\text{Block}_{\alpha}(\mathbf{X_i})-\mathbf{Y_i}||^2_2 \ , \nonumber
\\ & \mathbf{X_i} = 
\begin{cases}
\mathbf{Y_{i-1}},\text{Block}_i \in \{E_i^s\}  \\
\text{Cat}(\mathbf{Y_{i-1}},\mathbf{Skip_i}), \text{Block}_i \in \{D_i^s\} ,
\end{cases}
\nonumber
\end{align}
}

\noindent where $\mathcal{W}_{i}$ denotes the net parameters, $ \mathcal{A}_{i}$ denotes the architecture search space of $i$-th block, and $\alpha$ denotes the selected subnet. $\mathbf{Y_i}$ is the ground-truth feature map which is the output of the $i$-th block of the teacher. $\mathbf{X_i}$ denotes the input of the $i$-th block, which comes from the teacher. When it is an encoder block, $\mathbf{X_i}$ is the teacher output of the $(i-1)$-th block, or else it is the concatenation of the skip features and the teacher output of the $(i-1)$-th block. $\mathcal{L}_\text{train}$ is a L2 loss with mean reduction.

\begin{figure*}
    \centering
    \includegraphics[width=1.0\textwidth]{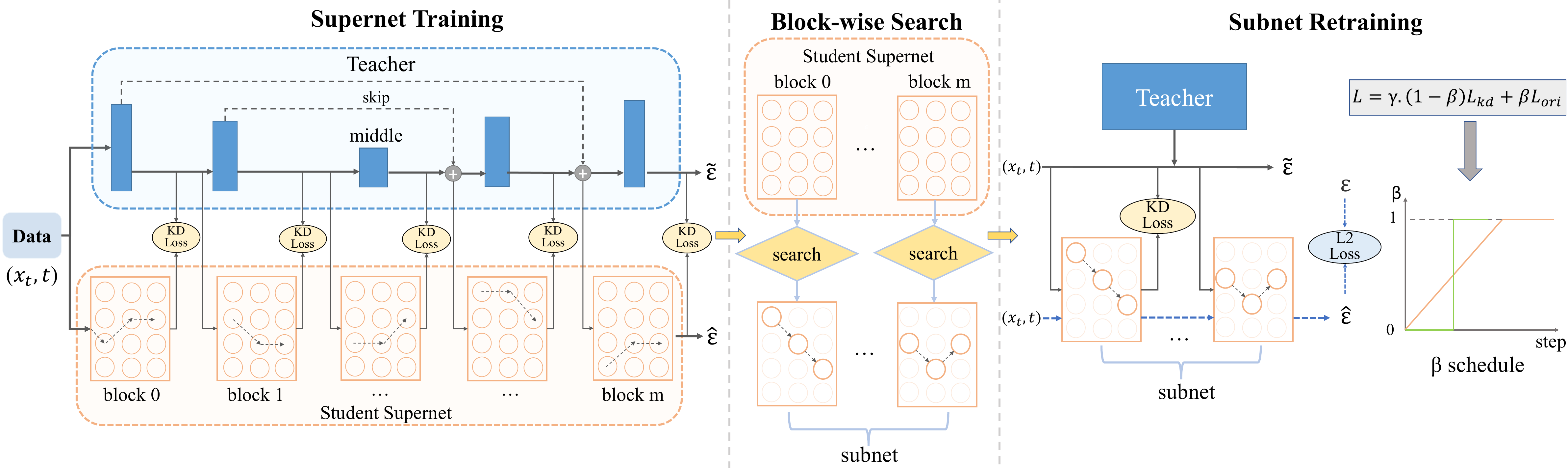}
    \caption{The illustration of our method DiffNAS, which consists of three stages, i.e., supernet training, block-wise search, and subnet retraining. In the supernet training, different blocks represent different levels of resolution. For each block, the intermediate feature maps of the teacher are used to supervise the training of the supernet. The input features of each student block also come from the previous outputs of the teacher. The mark '+' denotes the concatenation operation. Then we block-wisely search for the smallest subnet with better performance than the teacher, and concatenate the searched subnet in each block as the final net. Finally, we retrain the searched subnet with a dynamic joint loss. The black line denotes the path of the distillation loss, while the blue dashed line denotes the path of the original loss.}
    \label{framework}
\end{figure*}

The whole supernet training procedure is shown in the left part of Figure~\ref{framework}. Each block represents a resolution level. We use the random single-path strategy~\cite{guo2020single} for subnet sampling. It means that for each training step, we randomly select one operation at each layer and thus obtain a random architecture to be trained.
Note that for diffusion models, we observe that the $L2$ loss remains almost unchanged after training some epochs, even though further training continues to improve the generative performance. Therefore, we only need much fewer epochs for NAS supernet training compared to training a standalone diffusion model from scratch.

\subsection{Block-wisely Search for the Smallest Student}
After supernet training, we conduct the evaluation on the training dataset. We search for the model that has the least computation cost with on-par or even better performance than the teacher. Previous works~\cite{li2020block} use a global strategy to search for the appropriate subnets. Since different blocks have different scales of feature values, they use a relative loss such as the relative $L1$ loss to normalize the loss values of different blocks. The relative $L1$ loss is to divide the original $L1$ loss by the standard deviation among all elements of the ground-truth feature map.
Then they use the summation of these losses of each block as the final loss to evaluate the subnet performance. 
However, we observe that the relative loss also leads to unfairness, because some blocks have much larger losses than others.
It means that blocks with much larger losses dominate the final loss and thus dominate the search process. In this case, the summed loss is meaningless. Therefore, instead of the global search requiring the balance of losses, we propose to block-wisely select the subnet. Specifically, 
we use the subnet which has the same architecture as the teacher as the baseline. Let $\alpha_{base}$ denote it. For each block, we select the smallest subnet that has a better evaluation loss compared to $\alpha_{base}$. 
Finally, we obtain the complete subnet by concatenating the searched architectures of each block. 
For each block, the search strategy is shown as follows:

{
\begin{align}
\ \qquad &\alpha_{i}^{*}  = \underset{\alpha_{i} \in \mathcal{A}_{i}}{\arg \min } \ \text{Cost}(\alpha_{i}), \ i=0,1 \cdots 2n-1 , \nonumber \\
& \ s.t. \ \mathcal{L}_{\text {val}}\left(w_{i}, \alpha_{i} ; \mathbf{X_i}, \mathbf{Y_i}\right) \leq  r \mathcal{L}_{\text {val}}\left(w_{i}, \alpha_{base} ; \mathbf{X_i}, \mathbf{Y_i}\right) , \nonumber \\
&  \qquad  \text{where} \quad \mathcal{L}_{\text {val}} = ||\text{Block}_{\alpha}(\mathbf{X_i})-\mathbf{Y_i}||^2_2 \ , \nonumber
\end{align}
}

\noindent where Cost(.) represents the computation cost function with respect to the architecture and $\alpha_{i}$ denotes the architecture candidate of the $i$-th block. $\mathcal{L}_{\text{val}}$ denotes the validation loss. $r$ is a coefficient that can relax the constraint if we set it to a value larger than 1. 
Smaller subnets tend to have larger loss values, so we can increase $r$ slightly to obtain a sequence of smaller subnets. In this way, we can control the compression rate. 
Note that in our experiments, after retraining with our proposed strategy, the subnet with a $r$ slightly greater than 1 (e.g., 1.02) can also have better performance than the teacher.
The final optimal subnet is the concatenation of $\alpha_{i}^{*}$ of all blocks. Benefiting from the reduced search space brought by block-wise NAS, we can exhaustively traverse and evaluate all subnets in each block.
 To reduce the time costs, the evaluation for each subnet only computes several batches rather than the whole dataset, which largely speeds up the evaluation process. The experiments indicate this partial evaluation can achieve good enough results.
The Algorithm~\ref{algorithm:1} illustrates the whole search process, where ChooseBase denotes the function that chooses the subnet with the same architecture as the teacher.

\begin{algorithm}[h]
\DontPrintSemicolon
  \SetAlgoLined
  \KwIn {Supernet, Cost function, block input $\mathbf{X}$, ground-truth $\mathbf{Y}$, release coefficient $r$}
  \KwOut {The best lightweight model}
  \textbf{Define} SearchBlock(i):{ \;
    \qquad BaseNet = ChooseBase(Supernet[i]) \;
    \qquad $\text{loss}_\text{base}$ = $||\text{BaseNet}(\mathbf{X_i}) -\mathbf{Y_i} ||_2^2$ \;
    \qquad $\text{loss}_\text{min}$ =$\text{cost}_\text{min}$ = INF \;
     \qquad $\text{net}_\text{optimal}$ = None \;

    \qquad \For{Subnet in Supernet[i]}{
     \qquad $\mathbf{\hat{Y}_i}$ = Subnet($\mathbf{X_i}$) \;
     \qquad  loss = $||\mathbf{\hat{Y}_i}-\mathbf{Y_i}||^2_2$ \;
      \qquad cost = Cost(Subnet) \;
      \qquad   \If{$\text{loss} \leq r \text{loss}_\text{base}$}{
         \qquad  \If{$ cost < {cost}_{min} $}{
                 \qquad $\text{cost}_\text{min}$ = cost \;
                 \qquad $\text{loss}_\text{min}$ = loss  \;
                  \qquad $\text{net}_\text{optimal}$= Subnet \;
            \ \uElseIf{$ cost == {cost}_{min} $}{
                 \ \ \  \uIf{$loss < {loss}_{min}$}{
                     \ \  $\text{loss}_\text{min}$ = loss \;
                      \ \ $\text{net}_\text{optimal}$= Subnet  \;
                    }
                }
            }
       
            
        
        }
            
    }
  }
   Subnet = [] \;
  \For{$i \ in \ range(block\_num)$}{
      Subnet.append(SearchBlock(i)) \;
  }
   Subnet = Concatenate(Subnet) \;
  
  \textbf{Return} \ Subnet

  \caption{Search for the smallest Subnet.}
  \label{algorithm:1}
\end{algorithm}

\subsection{Subnet Retraining with Joint Dynamic Loss}

When retraining the searched subnet from scratch, we observe that only using the noise prediction leads to poor performance. 
In supernet training and evaluation, We use the distance between the student and the teacher to measure the quality of subnets. In other words, we actually measure the ability to imitate the teacher. Therefore, retraining the selected best student with the guidance of the teacher is important, which maintains the consistency between subnet evaluation and retraining. We propose a dynamic joint loss combining the distillation loss and the noise prediction loss, formulated as follows:

\begin{align}
& \mathcal{L}_{\text{retrain}} = \gamma \cdot (1-\beta)\mathcal{L}_\text{dis} + \beta \mathcal{L}_\text{ori} \ ,  \nonumber \\ 
& \mathcal{L}_\text{ori}=||\epsilon_{\theta}\left(\alpha_{t} x+\sigma_{t} \epsilon, t\right)-\epsilon||_2^2 \ ,  \nonumber \\
& \mathcal{L}_\text{dis} = \sum_{i}\mathcal{L}_{\text {train}} =\sum_i ||\text{Block}_i(\mathbf{X_i})-\mathbf{Y_i}||^2_2 \ ,  \nonumber
\end{align}

\noindent where $\beta$ is the weight that balances the distillation loss and the noise prediction loss. $\beta$ is set to 0 at the beginning and finally equals to 1. We use a linear or a step function to adjust $\beta$. The experiments show both functions are effective, and we recommend the step function as the first choice to save time.
$\mathcal{L}_\text{ori}$ is computed by passing $\alpha_t x +\sigma_t \epsilon$ and $t$ through all the layers, while $\mathcal{L}_\text{dis}$ is computed block-wisely whose inputs come from the teacher just like the NAS training.
This means in the initial training, we mainly use the guidance of the teacher. The guidance can provide more informative objectives, since the intermediate feature maps provide information on different abstraction levels and shorten the paths of gradient propagation. This can be regarded as a warm-up stage where the student quickly learns the knowledge from the teacher. In the later training, the student needs to be refined, so $\mathcal{L}_{ori}$ is more important since it directly determines the accuracy of noise prediction. We also use a parameter $\gamma$ to balance the learning rates of $\mathcal{L}_\text{dis}$ and $\mathcal{L}_\text{ori}$.

\begin{table*}[h]
\centering
\resizebox{0.6\linewidth}{!}{
\begin{tabular}{ccccccc}
\hline
Model & MACs (G)  & ddim-50 & ddim-100 & dpm-20 & dpm-25 \\ \hline
Teacher & 6.05  & 4.75 &  4.25  & 6.41 & 4.18 \\
S1(linear) & 4.18(-31\%) & 4.66 & 4.32 & 5.08 & 4.12 \\ 
S2(linear) & 4.51(-26\%) & \textbf{4.56} & \textbf{4.17}  & 4.69 & \textbf{3.92} \\
S2(step) & 4.51(-26\%) & 4.59 & 4.20  & 5.29 & 3.97 
\\ \hline
Random(linear) & 4.527(-26\%) & 5.41 & 4.85 &  \textbf{4.62} & 3.95 \\
Random(step) & 4.468(-26\%) & 4.69 & 4.33 &  5.65 & 4.21 \\
\hline
No-Dis-S2 & 4.510(-26\%) & 5.24 & 4.81 &  5.10 & 4.02 \\
Fixed-Loss-S1 & 4.175(-31\%) & 9.56 & - &  - & 11.97 
\\ \hline
\end{tabular}
}
\caption{CIFAR-10 image generation with pixel-wise model. 
}
\label{table: cifar}
\end{table*}

\begin{table}[htbp]
\centering
\resizebox{0.95\linewidth}{!}{
\begin{tabular}{ccccc}
\hline
Model & MACs (G) &  dpm-20 & dpm-25 & pndm-50
\\ \hline
Teacher & 23.88  & 4.46 &  4.33  & 5.66 \\
S1(linear) & 17.98(-25\%) &  4.16 & 4.34 &  \textbf{5.47} \\
S1(step) & 17.98(- 25\%) &  \textbf{3.73} & \textbf{4.03} &  5.53

\\  \hline
\end{tabular}
}
\caption{CelebA image generation with pixel-wise model.}
\label{table: celeba}
\end{table}

\begin{table}[htbp]
\centering
\resizebox{0.95\linewidth}{!}{
\begin{tabular}{cccccc}
\hline
Model & MACs (G) &  dpm-20 & dpm-25 & pndm-50
\\ \hline
Teacher & 62.05 & 13.61 &  10.82  & \textbf{12.86} \\
S1(step) & 46.38(-25\%) &  \textbf{12.34} & \textbf{10.51} &  13.93 \\  \hline
\end{tabular}
}
\caption{LSUN-church 128x128 image generation with pixel-wise model.}
\label{table: church}
\end{table}


\begin{table}[htbp]
\centering
\resizebox{1.05\linewidth}{!}{
\begin{tabular}{cccccc}
\hline
Dataset & Model & MACs (G) &  Params (M) &   FID & sFID 
\\ \hline
LSUN-church & Teacher & 18.64 & 294  &  5.26 & 24.85 \\
 & S1(step) & 10.12(-46\%)& 123(-58\%) &  5.75 & 25.34   \\  \hline
 CelebA-HQ & Teacher & 96.02 & 273  &  7.85 & 16.75  \\
 & S1(step) & 53.06(-45\%)& 149(-46\%) &  8.49 & 17.97   \\  \hline
\end{tabular}
}
\caption{LSUN-church 256x256 and CelebA-HQ 256x256 generation with latent diffusion models. Use DDIM sampler of 50 steps.
}
\label{table: ldm-celehq-church}
\end{table}

\section{Experiments}

\subsection{Implementation Details}
\textbf{Search space.} For simplicity, we only search the convolution kernel size of Resnet blocks which accounts for 
the majority of the total computation. 
Our method can also be suitable for larger search spaces such as including layers and channel numbers.
The standard UNet models use 3x3 convolution, so we take kernel sizes of \{1, 3, 5\} as search space. We construct a supernet for each block with each layer consisting of these three options.

\textbf{Supernet training.} 
For pixel-wise models, on CIFAR-10 and CelebA, we use the checkpoints provided by the DDPM and DDIM repositories respectively as the teacher. On LSUN-church, we train our own baseline model as the teacher. For latent models, we use the checkpoints provided by the official repository~\cite{rombach2022high}. The supernet training for different blocks can be done in parallel to speed up, because the input of each block comes from the teacher rather than the previous block. We find that the $L2$ loss tends to be unchanged after some epochs of training, even though further training continues to improve model performance. Therefore, we only need to train the supernet for much fewer epochs than training a baseline from scratch, and then we can do the evaluation. For example, training a teacher model for LSUN-church needs nearly 450,000 steps, while training the supernet only needs 1,750 steps for each block. When evaluating each subnet, we find computing for several batches is enough which can save much time.

\textbf{Subnet retraining.} At the beginning of subnet retraining, we set $\beta$ to 0, which means we only use the distillation loss. In this stage, the distillation loss transfers knowledge from the teacher to the subnet. Then we increase $\beta$ to assign a larger weight to the original noise prediction loss. Let $\beta$-$steps$ denote the number of training steps when $\beta$ increases to 1.
We test both the linear and the step functions. The linear function means $\beta$ increases from 0 to 1 linearly.
The step function means $\beta$ abruptly changes to 1 at $\beta$-$steps$.
We test these two functions on CIFAR-10 and CelebA.
The experiments show that compared to the linear strategy, the step strategy performs similarly on CIFAR-10 and noticeably better on CelebA. Both of them surpass the teacher and the ablation models.
we recommend the step function as the preferred choice since it is more time-efficient.
We observe that the training is not sensitive to the value of $\gamma$.

\begin{figure}
    \centering
    \includegraphics[width=0.40\textwidth]{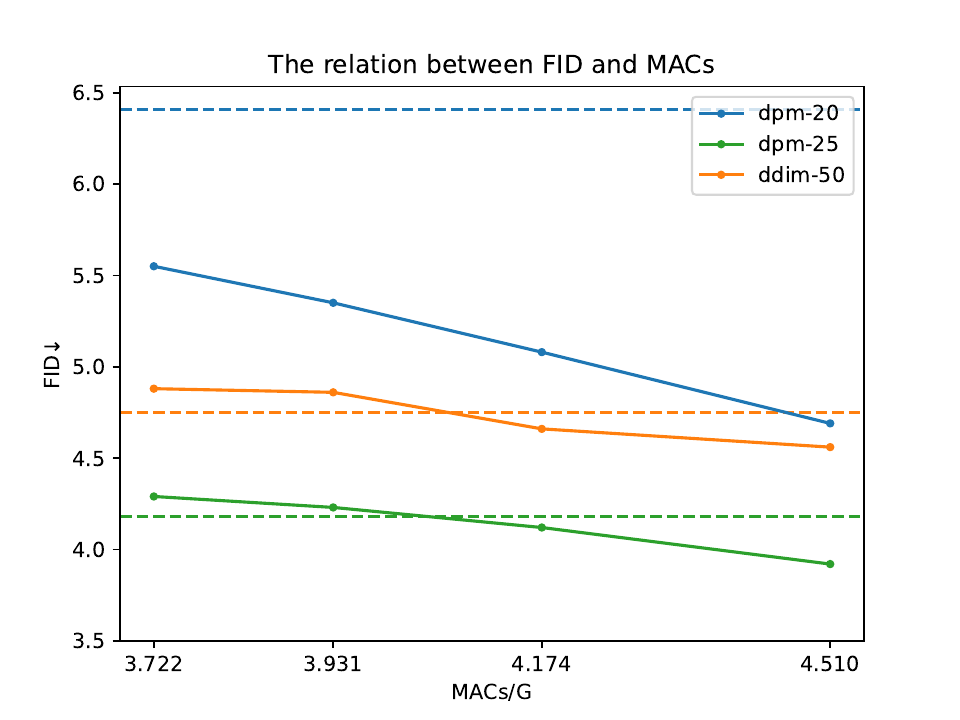}
    \caption{The curves of FID on CIFAR-10 with respect to MACs. 
    The dashed lines denote the FID of the teacher. Different colors represent different samplers.}
    \label{fig: fid-macs}
\end{figure}

\subsection{Results}

We use the most frequently-use metric FID~\cite{heusel2017gans} to measure the quality of image generation. We also compute sFID for latent diffusion models to better measure spatial similarity for high-resolution images. We test the classic sampler DDIM~\cite{song2020denoising} as well as advanced fast samplers, DPM-Solver~\cite{ludpm} and PNDM~\cite{liupseudo}. We use MACs (Multiply-Accumulate Operations) reduction to measure the compression ratio of computational cost. More details about evaluation settings can be found in the Appendix (\cref{sec: eval settings}).

The results on pixel-wise diffusion models. Table~\ref{table: cifar} shows the FID of our lightweight models and the teacher on CIFAR-10.
S1 is the searched model corresponding to $r=1.02$, while S2 corresponds to $r=1.00$. ``Random'' is a randomly selected architecture, while ``No-Dis-S2'' is a S2 model that is retrained without the joint loss. ``Fixed-Loss-S1'' is a S1 model trained with fixed rather than dynamic $\beta$.
we will discuss these three models in the ablation study.
The results show that S2 with the linear function
significantly outperforms the teacher under all samplers, and S1 with the linear function also outperforms the teacher except for ddim-100. We also find that the S2 with the step function achieves a very similar FID to the S2 with the linear function, except for dpm-20. Also, the S2 (step) noticeably outperforms the teacher.
As for compression ratio, S1 achieves a 31\% MACs reduction, while S2 achieves a 26\% reduction.
Table~\ref{table: celeba} shows the FID measurement on CelebA. S1 corresponds to $r=1.00$. Both the linear and the step function achieve better FID than the teacher. The step one performs better than the linear one. S1 achieves a 25\% MACs reduction. 
Table~\ref{table: church} shows the FID measurement on LSUN-church 128x128. S1 corresponds to $r=1.03$. The S1 with the step function
performs better than the teacher in general.

The results on latent diffusion models. Our method performs better on latent diffusion models, which achieve about 50\% MACs reduction and also Parameter reduction. Table~\ref{table: ldm-celehq-church} shows the results on LSUN-church 256x256 and CelebA-HQ with the latent diffusion models. Our compressed model has comparable performance to the teacher with significant MACs reduction (46\% and 45\%) and Parameter reduction (58\% and 46\%).


As mentioned earlier, we can obtain a series of search models with different compression ratios by slightly adjusting $r$. Smaller subnets tend to have larger evaluation losses, and vice versa. We test a series of models with different MACs on CIFAR-10.
Figure~\ref{fig: fid-macs} shows the relations between FID and subnet MACs on CIFAR-10.
The smaller MACs corresponds to a searched model under a larger $r$. The results indicate that the performance of subnets increases with respect to MACs, which accords with the intuition that more complex models usually have better performance.

\begin{figure*}[htbp]
    \centering
    \includegraphics[width=0.80\textwidth]{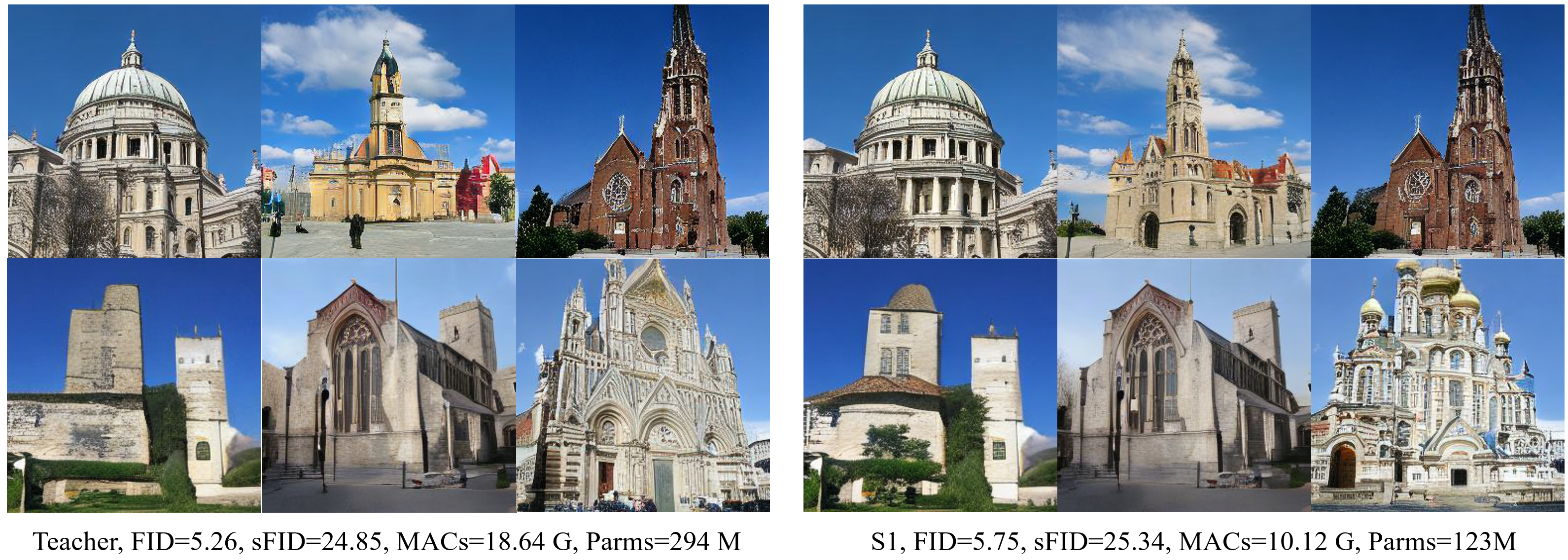}
    \caption{The generated images samples on LSUN-church 256 of our lightweight model and the teacher. The sampling strategy is DDIM with 50 steps.}
    \label{fig: ldm-church-grid}
\end{figure*}

\begin{figure*}[htbp]
    \centering
    \includegraphics[width=0.80\textwidth]{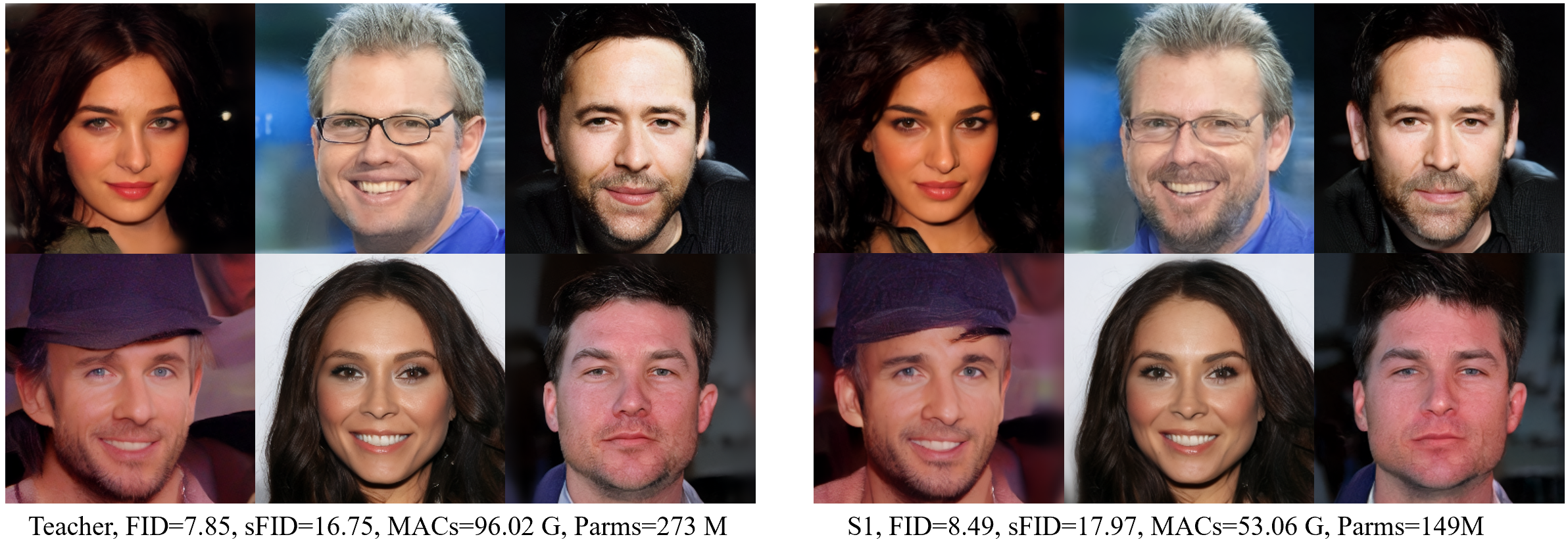}
    \caption{The generated images samples on CelebA-HQ 256 of our lightweight model and the teacher. The sampling strategy is DDIM with 50 steps.}
    \label{fig: ldm-celehq-grid}
\end{figure*}

Figure~\ref{fig: ldm-church-grid} shows the generated images of the original and our compressed latent models on LSUN-church 256x256. Figure~\ref{fig: ldm-celehq-grid} shows the results on CelebA-HQ 256x256. These qualitative results show there is no visual difference between our compressed model and the baseline. Samples of pixel-wise models can be found in the Appendix.

\subsection{Ablation Study}

\textbf{The effectiveness of NAS.} Previous works~\cite{yangevaluation,yuevaluating} suggest that a number of NAS methods are no better than or struggle to outperform
a random architecture. So it is important to prove that our method can surpass the random architecture. On CIFAR-10, We randomly select a model whose MACs nearly equals to S2 and train it with the same setting as S2. Table~\ref{table: cifar} shows the comparison between the random models and S2. 
The table shows that S2 (linear) outperforms Random (linear) significantly except for a similar FID at dpm-20. S2 (step) noticeably outperforms Random (step) under all samplers. These results prove the effectiveness of our DiffNAS, i.e., DiffNAS really obtains a noticeably better architecture than the random one.

\noindent \textbf{The effectiveness of the dynamic joint loss.} As mentioned before, we propose a joint loss to maintain the consistency between supernet training and  subnet retraining, provide guidance, and shorten paths of gradient propagation. On CIFAR-10, we train a model referred to as ``No-Dis-S2`` that represents a S2 model retrained without the distillation loss. 
The results in Table~\ref{table: cifar} show that both the S2 with the linear joint loss and the S2 with the step function noticeably outperform ``No-Dis-S2``. This proves the necessity of $\mathcal{L}_\text{dis}$.
Further, in order to prove the necessity to dynamically adjust the weight $\beta$, we conduct another experiment where we assign the same fixed weight to $\mathcal{L}_\text{ori}$ and $\mathcal{L}_\text{dis}$, which is denoted by Fix-Loss-S1 in Table~\ref{table: cifar}. The results show the model has very poor performance, which proves the necessity to reduce the weight of $\mathcal{L}_\text{dis}$ in the later steps of training.

\noindent \textbf{The effectiveness of block-wise search.} As mentioned before, our method proposes to block-wisely search for the most appropriate subnets to avoid the unfairness among blocks of the global search strategy adopted by previous works. This unfairness is due to the large differences of loss values among different blocks, even after being balanced. For example, we observe that some blocks have loss values about 0.50, while some other blocks have loss values about 0.02. Therefore, the global search strategy which summarizes the losses of all blocks will be dominated by the blocks that have large losses and ignore the blocks with small losses. To prove this proposition, we conduct an experiment that uses an evolutionary algorithm to globally search for the smallest subnet, on the CIFAR-10 dataset. We use the relative $L2$ loss shown below to balance the loss values among blocks: 

{
\begin{align}
 \mathcal{L}_{\text {val}} = \frac{ ||\text{Block}_i(\mathbf{X_i})-\mathbf{Y_i}||^2_2 }{\text{var}(\mathbf{Y_i})} \ , \nonumber
\end{align}
}

\noindent where $\text{var}(\mathbf{Y_i})$ denotes the variance of the ground-truth. After performing the evolutionary algorithm for 30 epochs, we obtain a subnet that has a better evaluation loss than the teacher. However, this subnet has an obviously poor performance compared to the teacher performance in Table \ref{table: cifar}, with the FID of 5.08, 8.23, and 5.74 on ddim-50, dpm-20, and dpm-25 respectively. This proves that our block-wise search is much more effective than the global one.

\section{Conclusion}

To reduce the computational cost of diffusion models, we propose DiffNAS which automatically searches for the smallest architecture while maintaining the performance as much as possible. DiffNAS contains a novel block-wise search strategy and a novel retraining strategy with a dynamic joint loss.
The block-wise search strategy can avoid the unfairness of loss values among different blocks. 
The dynamic joint loss can maintain the consistency between supernet training and subnet retraining, and transfer teacher knowledge to improve performance.
Empirical results show that our method achieves about 30\% MACs reduction on pixel-wise models, as well as about 50\% MACs and Parameters reduction on latent diffusion models, while maintaining on-par or even achieving better performance.

{
    \small
    \bibliographystyle{ieeenat_fullname}
    \bibliography{main}
}

\clearpage
\setcounter{page}{1}
\maketitlesupplementary

\section{Extended Experimental Settings}

\subsection{Datasets}
To demonstrate the generalization of our method, we conduct experiments on several datasets with different image resolutions.

\textbf{CIFAR-10~\cite{krizhevsky2009learning}.} CIFAR-10 is a popular image classification dataset that contains 60,000 32x32 
RGB images from 10 different classes. CIFAR-10 is a popular baseline for computer vision tasks such as image generation.

\textbf{CelebA~\cite{liu2015faceattributes} and CelebA-HQ~\cite{karras2017progressive}.} CelebA is a large-scale face attributes dataset with 202,599 celebrity images. CelebA-HQ is a higher-resolution dataset developed from CelebA. They are also commonly used in image generation tasks.

\textbf{LSUN-church~\cite{yu2015lsun}.} LSUN-church is part of the Large-Scale Scene Understanding (LSUN) datasets, consisting of 126,527 outdoor church images.
LSUN-church covers a wide range of church architectures and styles, making it diverse and challenging. LSUN-church is a frequently-used dataset in image generation to measure the performance of generative models.

\subsection{Evaluation Settings}
\label{sec: eval settings}

We use FID and sFID to measure the quality of generative models. FID is an important metric used in image generation tasks to measure the similarity between generated images and real images. A lower FID indicates higher similarity and always means better quality of generative models. sFID~\cite{nash2021generating} is another frequently-used metric, a variant of FID, which can better measure spatial similarity.
For all pixel-wise models, we generate 50,000 samples and compute the FID score between these generated images and the whole training dataset. For latent diffusion models, we generate 10,000 samples and calculate both FID and sFID. For the measurement of MACs and parameter amount, we use the \textit{THOP} package.


\section{Computational Costs of Our Method}
\label{sec: cost analysis}

Our method only requires a little extra time cost and no extra memory cost compared to training a baseline. Specifically:
\begin{itemize}
\item The supernet training stage is fast. As reported in the paper, we only need a few steps to train each block, e.g., 1,750 steps using pixel-wise diffusion and 2,000 steps using latent diffusion, on LSUN-Church. Besides, because we only need to train a block once a time, the memory cost is much less than training a baseline. We cost about 6 hours on two A100 GPUs when testing on Church 256x256 with the latent diffusion model.

\item The search stage is fast. As discussed in the paper, we evaluate each subnet in each block with several batches, so the stage is fast. Besides, this stage doesn't need training, so it is also memory-efficient. We cost about 1.2 hours in a single A100 GPU when testing on Church with the latent diffusion model

\item The retraining stage also does not require extra memory cost. Our method retrains the compressed subnet, so it needs less GPU memory cost than training a baseline. 
We cost about 20G memory (batch size=32 for each GPU) when testing on Church with the latent diffusion model, while training a baseline costs about 25G memory. 
The stage only needs a little extra time cost to do teacher inference at the beginning when the beta is less than 1. Moreover, We also find the number of converged epochs is less than the baseline. 

\end{itemize}

\section{Qualitative Results of Pixel-wise Diffusion Models}

\begin{figure*}[t]
    \centering
    \includegraphics[width=0.95\textwidth]{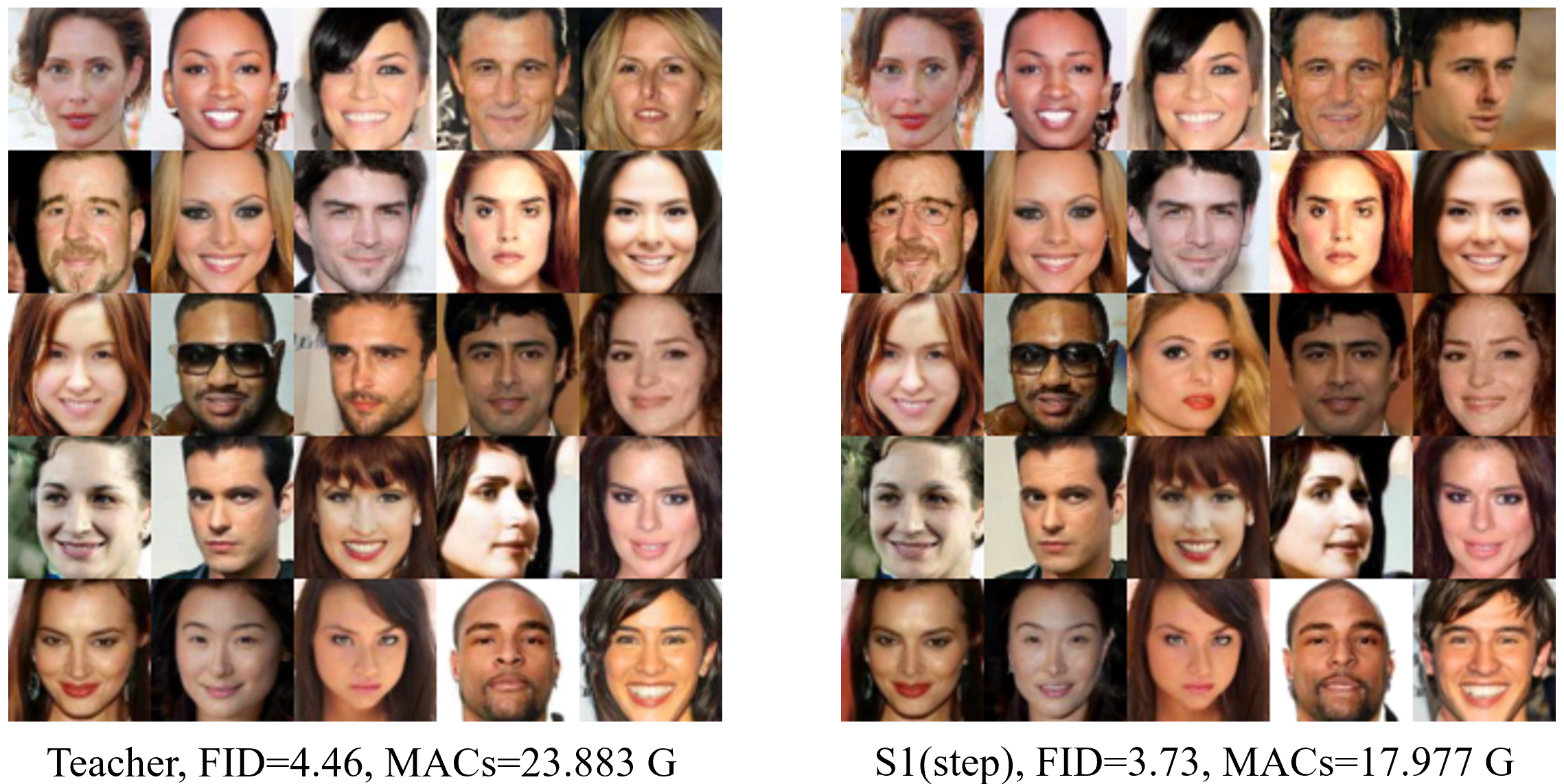}
    \caption{The generated images on CelebA of our lightweight model and the teacher. The sampling strategy is a DPM-Solver of 20 steps.}
    \label{fig: cele_grid}
\end{figure*}

\begin{figure*}[t]
    \centering
    \includegraphics[width=0.95\textwidth]{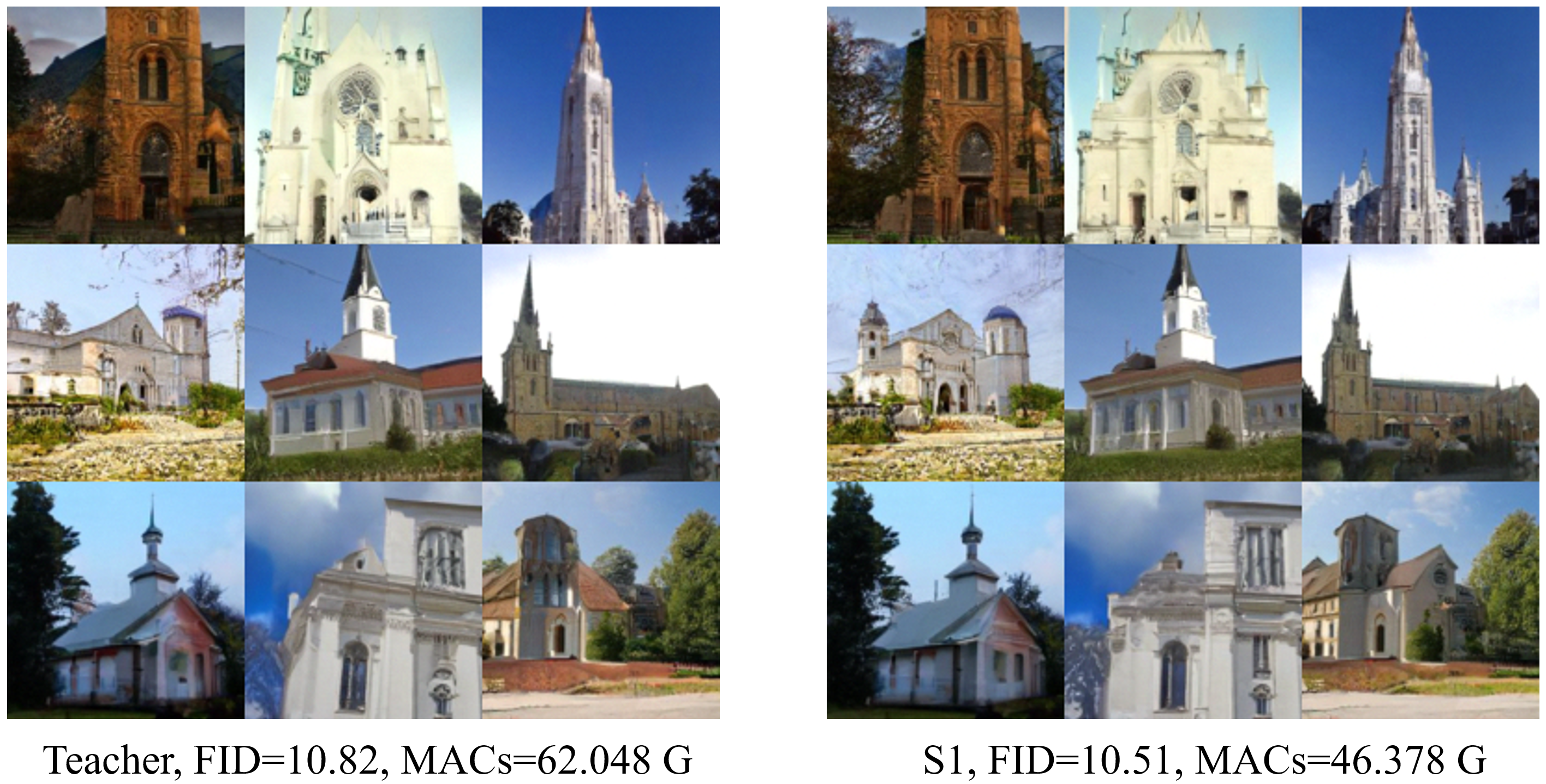}
    \caption{The generated images on LSUN-church 128. The sampling strategy is a DPM-Solver of 25 steps.}
    \label{fig: church_grid}
\end{figure*}

We show the generated images of pixel-wise diffusion models on CelebA 64x64 and LSUN-Church 128x128, i.e., Figure~\ref{fig: cele_grid} and \ref{fig: church_grid}. These qualitative results show there is no visual difference between our compressed model and the baseline.



\end{document}